\pdfoutput=1

\documentclass[11pt]{article}

\usepackage[final]{acl}

\usepackage{times}
\usepackage{latexsym}

\usepackage[T1]{fontenc}

\usepackage[utf8]{inputenc}

\usepackage{microtype}

\usepackage{inconsolata}

\usepackage{graphicx}
\usepackage{makecell}
\usepackage{multirow}
\usepackage{caption}
\usepackage{todonotes}

\usepackage{enumitem}

\title{MultiCW: A Large-Scale Balanced Benchmark Dataset for Training Robust Check-Worthiness Detection Models}

\author{
    Martin Hyben$^\dagger$,
    Sebastian Kula$^\dagger$$^{\spadesuit}$,
    Jan Cegin$^\dagger$,
    Jakub Simko$^\dagger$,
    Ivan Srba$^\dagger$,
    Robert Moro$^\dagger$
    \\
    $^\dagger$ Kempelen Institute of Intelligent Technologies, Bratislava, Slovakia\\
    $^{\spadesuit}$ West Pomeranian University of Technology in Szczecin, Szczecin, Poland \\
    \texttt{\{name.surname\}}@kinit.sk\\ 
}

\begin{document}
\maketitle
\begin{abstract}
    Large Language Models (LLMs) are beginning to reshape how media professionals verify information, yet automated support for detecting check-worthy claims—a key step in the fact-checking process—remains limited. We introduce the Multi-Check-Worthy (MultiCW) dataset, a balanced multilingual benchmark for check-worthy claim detection spanning 16 languages, 7 topical domains, and 2 writing styles. It consists of 123,722 samples, evenly distributed between noisy (informal) and structured (formal) texts, with balanced representation of check-worthy and non-check-worthy classes across all languages. To probe robustness, we also introduce an equally balanced out-of-distribution evaluation set of 27,761 samples in 4 additional languages.
    To provide baselines, we benchmark 3 common fine-tuned multilingual transformers against a diverse set of 15 commercial and open LLMs under zero-shot settings. Our findings show that fine-tuned models consistently outperform zero-shot LLMs on claim classification and show strong out-of-distribution generalization across languages, domains, and styles. MultiCW provides a rigorous multilingual resource for advancing automated fact-checking and enables systematic comparisons between fine-tuned models and cutting-edge LLMs on the check-worthy claim detection task.
\end{abstract}

\section{Introduction}
The work of media professionals, such as fact-checkers and journalists, involves processing large volumes of textual information daily to identify key pieces of information~\cite{micallef_true_2022}. This content can originate from diverse sources, ranging from well-structured documents (e.g., news articles, press releases) to highly noisy formats (e.g., social media posts, interview transcripts). A major challenge, apart from the large amount of texts, arises from the linguistic diversity of these sources: media professionals often encounter content in languages they do not speak, leaving them with no option but to rely on laborious and inefficient translations. In this demanding environment, automated systems that highlight the most relevant and verifiable information can significantly reduce manual effort~\cite{10.1145/3764592}.

One promising direction is the \textit{automated detection of check-worthy (CW) claims}, defined as information units that are sufficiently important, verifiable, or impactful to warrant further examination~\cite{alam-etal-2021-fighting-covid, srba2024surveyautomaticcredibilityassessment, panchendrajan2024claimdetection}. We follow prior work in defining CW claims as information units that best represent the core informational value of a text and are central to fact-checkers’ workflows \cite{DBLP:journals/corr/abs-1808-05542}. However, since the notion of check-worthiness varies across sources, we propose a unified operationalization of its definition, which is discussed in detail in Section \ref{sec:cw_definition}.

\begin{table}[ht]
\centering
\small
\begin{tabular}{|p{3.5cm}|p{3.5cm}|}
\hline
\textbf{Check-worthy Claim} & \textbf{Non-check-worthy Claim} \\
\hline
"The COVID-19 vaccine alters your DNA." & "I feel much better after getting vaccinated." \\
\hline
\end{tabular}
\caption{Examples of check-worthy and non-check-worthy claims.}
\label{tab:claim_examples}
\end{table}

Table \ref{tab:claim_examples} illustrates the difference between a check-worthy and a non-check-worthy claim. The CW claim in the table satisfies the criteria for being check‑worthy: significance, impact, and (potentially) source reliability. In contrast, the non‑CW claim is subjective, trivial, and lacks impact, thereby meeting the criteria for being non‑check‑worthy.

Despite growing interest in this task, existing datasets for CW detection remain limited: they are often constrained to English ~\cite{gupta-etal-2021-lesa}, to narrow domains (e.g., COVID-19) ~\cite{Savchev:CLEF2022}, or to specific styles such as headlines. These limitations hinder the ability of automated systems to generalize to multilingual, multi-topic, or user-generated content.

To address these gaps, we introduce \textit{Multi Check-Worthy (MultiCW) dataset}\footnote{The dataset is available at: 
\url{https://zenodo.org/records/17482958}}, a large-scale multilingual resource designed to benchmark CW detection across languages, topics, and writing styles. MultiCW integrates high-quality claims from multiple sources -- existing datasets, additional samples from Wikipedia and translations of English claims from predominantly English datasets. The result is a balanced dataset spanning 16 languages (Table \ref{tab:dataset_balanced}), 7 topics (health, politics, environment, science, sport, entertainment, history), and 2 distinct writing styles: noisy text, characterized by informal phrasing, spelling errors, slang, and conversational tone; and structured text, written in a formal tone with clear syntax, proper grammar, and well‑organized sentences.

The purpose of MultiCW is to serve as a currently missing benchmark dataset for evaluating solutions to check-worthy claim detection\footnote{A repository with the full code, preprocessing scripts, and dataset reconstruction pipeline (including splits) is available at \url{https://github.com/kinit-sk/MultiCW}}. To this end, we also establish strong baselines by training three multilingual Transformer-based models (Table \ref{tab:transformer_multicw}) and by prompting a large variety of open and commercial large language models in zero-shot settings (Table \ref{tab:llm_results}). Our experiments demonstrate that fine-tuned models consistently outperform zero-shot LLMs, highlighting both the challenge of the task and the utility of MultiCW as a benchmark.

In addition to in-domain benchmarks, we further evaluate the generalization capabilities of fine-tuned models on a separate out-of-distribution dataset of 27,761 claims in 4 additional languages (Table \ref{tab:ood_balanced}). This experiment provides insights into how well models trained on MultiCW transfer to unseen languages and domains.

\section{Related Work}
\label{sec:related_work}

Large Language Models (LLMs) \citep{brown2020language} have garnered significant attention in recent years. Notable examples such as ChatGPT, Claude, Llama and Mistral have seen widespread adoption, with applications across various domains, including disinformation detection. 


 Several established approaches, as well as datasets are also focused on detecting check-worthy claims, particularly through fine-tuned language models. The detection of check-worthy documents and claims is a crucial task that has garnered significant attention from the human fact-checking community, whose expectations regarding the utility of AI-powered tools in facilitating their work have been well-documented in \citet{10.1145/3764592}. For instance, \citet{gupta-etal-2021-lesa} proposed the LESA framework for claim detection, utilizing BERT for semantic feature extraction and BiLSTM for linguistic features. While the model has the advantage of general applicability across different types of online content, its limitation lies in the utilized dataset, which is restricted to English and primarily focused on COVID-19-related data. \citet{DBLP:conf/emnlp/SundriyalKPA022} developed an approach for identifying claim spans in Twitter posts using DABERT, a variant of RoBERTa. However, this method’s applicability is constrained by the use of data from a single source (Twitter). 

\citet{kula2024multilingualmodelscheckworthysocial} presented an extensive study of models for detection of social media posts that contain verifiable factual claims and harmful claims. LLMs such as Alpaca-LoRA, llama-3.1 405b-instruct, and llama-3.1-70b-instruct were tested for verifiable factual claims and harmful claims detection tasks.


\citet{DBLP:conf/acl/GuptaWLX22} introduced a benchmark for verifiable claim detection in dialogues, presenting an analysis of three different methods: lexical overlap, Dialogue Natural Language Inference (DNLI), and a combination of both. Their study was also limited to English datasets. 

\citet{ni-etal-2024-afacta} introduced a reliable approach to facilitate the annotation of factual claims, harnessing the power of LLMs. This innovative framework offers the different reasoning paths as a method to enhance the accuracy and efficiency of claim annotation, thereby contributing significantly to the field of fact-checking.

Finally, \citet{atanasova2018overviewclef2018checkthatlab,elsayed2021overviewclef2019checkthatautomatic,DBLP:conf/clef/NakovBMAMCKZLSM20,DBLP:conf/clef/ShaarHHAHNKKAMB21,DBLP:conf/clef/NakovBMAMCKZLSM22,DBLP:conf/clef/NakovBMAMCKZLSM24} have reviewed methods and results from multiple shared tasks, one of which focused on detecting check-worthy claims in tweets. Although the datasets included six languages and focused on COVID-19-related content, most models were trained on single-language datasets, with only a few models trained in a multilingual setting. Notably, mT5, AraBERT, and GPT-3 achieved top results depending on the language and dataset.

We can conclude that while significant progress has been made in check-worthy claim detection, many of these efforts are constrained by the use of limited data sources (e.g., COVID-19-related content, single platforms such as Twitter) and language (primarily English). Our study seeks to mitigate the limitations of the current methods and contribute to the development of more comprehensive solutions.

\section{Check-Worthiness Definition}
\label{sec:cw_definition}

The notion of what constitutes a check-worthy (CW) claim is not always clear-cut. Rather than a strict binary distinction, check-worthiness can be seen as a continuum: some claims are highly significant, others moderately significant, while many are trivial and thus not check-worthy. However, for the purposes of consistent annotation and dataset construction, we adopt a practical binary definition based on prior work~\cite{aarnes2024iaigroupcheckthat2024, ni2024afactaassistingannotationfactual, alam-etal-2021-fighting-covid} and our own criteria.

\textbf{Check-Worthy Claims.} We consider a claim to be check-worthy if it meets one or more of the following criteria:  
\begin{enumerate}[noitemsep, topsep=0pt]
    \item \textbf{Significance:} The claim carries implications for public policy, health, safety, or societal well-being.  
    \textit{Example: ``A new variant of COVID-19 has been detected in Europe.''}  
    \item \textbf{Controversy:} The claim is disputed or likely to spark public or expert debate.  
    \textit{Example: ``Global warming is a hoax.''}  
    \item \textbf{Impact:} The claim can influence public opinion, shape decisions, or alter behavior.  
    \textit{Example: ``The government has cut taxes for low-income households.''}  
    \item \textbf{Source Reliability:} The claim originates from a public figure, authority, or institution with wide influence.  
    \textit{Example: ``The President announced new sanctions.''}  
\end{enumerate}

\textbf{Non-Check-Worthy Claims.} Conversely, a claim is generally considered non-check-worthy if it falls into at least one of the following categories:  
\begin{enumerate}[noitemsep, topsep=0pt]
    \item \textbf{Subjectivity:} Purely subjective statements of opinion or taste.  
    \textit{Example: ``Chocolate ice cream is the best flavor.''}  
    \item \textbf{Triviality:} Inconsequential claims without broader implications.  
    \textit{Example: ``It rained in Paris yesterday.''}  
    \item \textbf{Common Knowledge:} Widely accepted facts that do not introduce new or contested information.  
    \textit{Example: ``The Earth orbits the Sun.''}  
    \item \textbf{Lack of Impact:} Claims with minimal influence or negligible consequences if false.  
    \textit{Example: ``The shop on 5th Street closes at 9 PM.''}  
\end{enumerate}

This working definition guided our dataset construction and annotation, ensuring consistent labeling of check-worthy and non-check-worthy claims across languages, topics, and styles.

\section{MultiCW Dataset Construction}
\label{sec:dataset_construction}

The MultiCW dataset is not a simple aggregation of prior resources, but a carefully curated collection designed to support robust multilingual CW detection. We integrate several existing datasets, the most prominent include: (i) CLEF-2022 \cite{DBLP:conf/clef/NakovBMAMCKZLSM22} and CLEF-2023 \cite{clef-checkthat:2023:task1}, which provide noisy claims with binary annotations across six languages; (ii) MultiClaim \cite{pikuliak-etal-2023-multilingual} and its extended version MultiClaim v2\footnote{\url{https://zenodo.org/records/15413169}}, which contribute structured fact-checking articles (check-worthy by definition) and social media claims in 39 languages; and (iii) Ru22Fact \cite{zeng-etal-2024-ru22fact}, a multilingual fact-checking dataset on the Russia-Ukraine conflict in 2022. 

Beyond merging these resources, we perform several steps to ensure balance and coverage. First, we harmonize the data across \textit{writing styles} (structured, noisy) and \textit{labels} (check-worthy vs.\ non-check-worthy). Next, we address under-represented languages and classes through targeted translations. Finally, we partition the dataset into an \textit{in-distribution} subset spanning 16 languages and an \textit{out-of-distribution} (OOD) subset covering 4 additional languages, enabling controlled evaluation of multilingual generalization. 

A detailed description of each source dataset and the harmonization process is provided in the following subsections and in Appendix~\ref{sec:appendix-datasets}.

\begin{table}[h]
\centering
\resizebox{0.74\linewidth}{!}{
\begin{tabular}{|c|r|r|r|}
\hline
\textbf{Language} &
\makecell{\textbf{Noisy} \\ \textbf{(class 0)}} &
\makecell{\textbf{Noisy} \\ \textbf{(class 1)}} &
\makecell{\textbf{Structured} \\ \textbf{(class 1)}} \\
\hline
ar & 5895 & 7084 & 21152 \\
bg & 2118 & 845 & 329 \\
bn & 0 & 1953 & 4118 \\
cs & 0 & 568 & 14559 \\
de & 0 & 3839 & 7673 \\
en & 20374 & 47395 & 152702 \\
es & 16402 & 24959 & 25733 \\
fr & 0 & 4980 & 6404 \\
hi & 0 & 5793 & 11040 \\
pl & 0 & 2487 & 8803 \\
pt & 0 & 5912 & 21512 \\
ru & 0 & 1149 & 5628 \\
sk & 0 & 504 & 20456 \\
tr & 3007 & 4660 & 12828 \\
uk & 0 & 29 & 3601 \\
zh & 0 & 559 & 3862 \\
\hline
\end{tabular}}
\caption{Statistics for the \textit{in-distribution} part of the \textit{unbalanced} MultiCW dataset.}
\label{tab:dataset_unbalanced}
\end{table}

\begin{table}[h]
\centering
\resizebox{0.74\linewidth}{!}{
\begin{tabular}{|c|r|r|r|}
\hline
\textbf{Language} &
\makecell{\textbf{Noisy} \\ \textbf{(class 0)}} &
\makecell{\textbf{Noisy} \\ \textbf{(class 1)}} &
\makecell{\textbf{Structured} \\ \textbf{(class 1)}} \\
\hline
it & 0 & 1482 & 3021 \\
mk & 0 & 1385 & 1123 \\
my & 0 & 1340 & 1297 \\
nl & 1090 & 1910 & 1227 \\
\hline
\end{tabular}}
\caption{Statistics for the \textit{out-of-distribution} part of the \textit{unbalanced} MultiCW dataset.}
\label{tab:ood_unbalanced}
\end{table}

\begin{table}[h]
\centering
\resizebox{0.95\linewidth}{!}{
\begin{tabular}{|c|r|r|r|r|}
\hline
\textbf{Language} &
\makecell{\textbf{Noisy} \\ \textbf{(class 0)}} &
\makecell{\textbf{Noisy} \\ \textbf{(class 1)}} &
\makecell{\textbf{Structured} \\ \textbf{(class 0)}} &
\makecell{\textbf{Structured} \\ \textbf{(class 1)}} \\
\hline
ar & 2000 & 1993 & 2000 & 2000 \\
bg & 2000 & 1993 & 1093 & 1093 \\
bn & 1964 & 1953 & 2000 & 2000 \\
cs & 1961 & 1992 & 2000 & 2000 \\
de & 1959 & 2000 & 2000 & 2000 \\
en & 2000 & 2000 & 2000 & 2000 \\
es & 2000 & 1997 & 2000 & 2000 \\
fr & 1962 & 2000 & 2000 & 2000 \\
hi & 1968 & 2000 & 2000 & 2000 \\
pl & 1963 & 2000 & 2000 & 2000 \\
pt & 1961 & 2000 & 2000 & 2000 \\
ru & 1969 & 1997 & 2000 & 2000 \\
sk & 1962 & 1986 & 2000 & 2000 \\
tr & 2000 & 1898 & 2000 & 2000 \\
uk & 1956 & 1986 & 2000 & 2000 \\
zh & 1960 & 1984 & 2000 & 2000 \\
\hline
\end{tabular}}
\caption{Statistics for the \textit{in-distribution} part of the \textit{balanced} MultiCW dataset.}
\label{tab:dataset_balanced}
\end{table}

\begin{table}[h]
\centering
\resizebox{0.95\linewidth}{!}{
\begin{tabular}{|c|r|r|r|r|}
\hline
\textbf{Language} &
\makecell{\textbf{Noisy} \\ \textbf{(class 0)}} &
\makecell{\textbf{Noisy} \\ \textbf{(class 1)}} &
\makecell{\textbf{Structured} \\ \textbf{(class 0)}} &
\makecell{\textbf{Structured} \\ \textbf{(class 1)}} \\
\hline
it & 2000 & 1482 & 2000 & 2000 \\
mk & 1999 & 1385 & 1123 & 1123 \\
my & 1999 & 1340 & 1297 & 1297 \\
nl & 1999 & 1910 & 1227 & 1227 \\
\hline
\end{tabular}}
\caption{Statistics for the \textit{out-of-distribution} part of the \textit{balanced} MultiCW dataset.}
\label{tab:ood_balanced}
\end{table}

\subsection{Balancing Strategy}
A key contribution of MultiCW is its careful balancing across \textbf{languages}, \textbf{writing styles}, and \textbf{classes}. The original datasets are highly skewed (e.g., Spanish dominates CLEF-2022 and CLEF-2023, English dominates MultiClaim), with class imbalances of up to 75\% non-check-worthy claims. Moreover, they do not provide any structured non-check-worthy samples. Table~\ref{tab:dataset_unbalanced} and Table~\ref{tab:ood_unbalanced} provide statistics on number of samples before balancing for the in-distribution and the out-of-distribution parts respectively. 

To address this high imbalance, we applied a two-step balancing strategy:  

\begin{itemize}
    \item \textbf{Writing style balance:} We first separated the data from the source datasets into two subsets based on the writing style: \emph{structured} and \emph{noisy}, taking into account the origin of the data (i.e., fact-checking articles were considered to be in structured writing style).
    \item \textbf{Class and language balance:} Within each writing-style subset, we enforced an equal number of check-worthy and non-check-worthy claims for each language. Over-represented languages were downsampled to 2000 samples, and under-represented ones were supplemented through translation-based and wikipedia sampling augmentation. To maintain the writing style of the augmented samples, we take into account the original annotations provided by the authors (i.e., the LESA dataset).
\end{itemize}

To enrich under-represented languages and domains, we applied two augmentation techniques:
\begin{itemize}
    \item \textbf{Translation:} For the noisy subset, we translated noisy and semi-noisy claims from the English-only LESA dataset into under-represented languages using a cloud-based MT system (DeepTranslate~\cite{baccouri2023deep}). For OOD dataset we translated the claims from ClaimBuster dataset~\cite{fatma_arslan_2020_claimbuster}. Prior to translation, the samples were filtered using the Gemma3 4B model to remove instances without valid claims. Gemma3 was selected due to its multilingual coverage and lightweight architecture, which ensured efficient processing. In addition, we translated all samples in the dataset into English to provide the original English samples as well as the cross-lingual consistency.
    \item \textbf{Wikipedia sampling:} To address the heavy imbalances in the \emph{structured} subset (specifically the MultiClaim dataset) with respect to both class distribution and language coverage, we extracted named entities from the unbalanced dataset using the GLiNER model \cite{zaratiana2023gliner} and retrieved the corresponding Wikipedia pages. The resulting sentences were included exclusively as non-check worthy as Wikipedia is considered a common knowledge. The extracted entity topics were also preserved in the dataset metadata. While natural class imbalance is realistic for deployment, a benchmark’s purpose is standardized, controlled evaluation. Thus, balancing the dataset is a deliberate and justified design choice to isolate model performance rather than reflect the incidental statistics of the source corpora.
\end{itemize}

Each data sample was labeled by its origin (\textit{manual}, \textit{translated}, \textit{wikipedia}) to maintain transparency.

\subsection{Quality Control}
To ensure quality, we applied multiple filtering steps: (i) removal of duplicates and empty strings, (ii) filtering of claims exceeding 5000 characters or containing no alphabetic characters, and (iii) manual spot-checks of a random subset. This process eliminated noise while preserving diversity.

We evaluated translations for three high-resource (English, German, French) and three low-resource languages (Czech, Slovak, Polish), covering both structured and noisy texts.

\textit{Structured texts} were translated with high fidelity across all languages. Errors were rare and mainly caused by overly literal translations that altered meaning (e.g., translating ``delete coronavirus'' instead of the intended ``release coronavirus''). For low-resource languages, 13/150 translations contained issues, primarily literal/contextual errors, along with 2 incorrect name translations and 3 cases of missing words. High-resource languages showed fewer issues (9/150), following the same pattern, with 1 incorrect name translation and 3 missing-word cases.

\textit{Noisy texts} posed greater challenges due to slang, abbreviations, informal language, and named entities (e.g., translating ``Judas Priest'' literally instead of as a band name). As many noisy samples were non-English, we evaluated translations in both directions. Low-resource languages exhibited 32/150 problematic translations, dominated by literal/contextual errors, with occasional incorrect name translations, missing words, and mistranslated abbreviations. High-resource languages showed fewer issues (17/150), including similar error types and a small number of partially missing translations.

Overall, structured texts translated reliably across languages, while noisy texts, especially in low-resource settings, were more susceptible to literal translations, entity handling errors, and omissions. Nevertheless, these inaccuracies did not significantly impact the CW claim detection task or dataset quality.

\subsection{Dataset Statistics}
\label{sec:dataset_statistics}
The final MultiCW dataset contains 123,722 claims spanning 16 languages, 7 topics, and 2 writing styles. Table~\ref{tab:dataset_balanced} presents the per-language and per-class distribution of this balanced dataset. In addition, we construct an out-of-distribution (OOD) evaluation set with 27,761 samples covering 4 other languages, with source datasets and balancing process similar to that of MultiCW. The 4 other languages were obtained by lowering initial threshold for minimal number of samples per language and class included in the unbalanced pool from 1500 for the MultiCW dataset to 1000 for OOD dataset. The OOD set serves as a generalization test by introducing new languages and altered topic distributions that were not seen during training. Table~\ref{tab:ood_balanced} summarizes its per-language and per-class distribution. Taken together, the two datasets comprise over 150k samples across 20 languages, balanced along writing style, and binary class labels. Unlike prior datasets, MultiCW is deliberately balanced across all three axes, enabling more robust evaluation of multilingual models. Dataset samples are almost equally distributed across languages, ranging from 5\% (Bengali) to 6.5\% (English) (see Figure~\ref{fig:lang_distribution_multicw} in Appendix A).

The dataset is divided into train, development, and test splits with the following ratio: 70:15:15, which constitute the split sizes of 
86,185 training samples, 18,387 validation samples, and 18,444 test samples. 
All splits are balanced across classes, languages, and writing styles. 
The separate out-of-distribution split containing 4 independent languages is of a similar size to the test set and contains 27,761 samples, allowing us to evaluate model robustness across different input conditions in a similar configuration as for in-domain experiments.

\begin{figure}[!ht]
    \centering
    \includegraphics[width=0.4\textwidth]{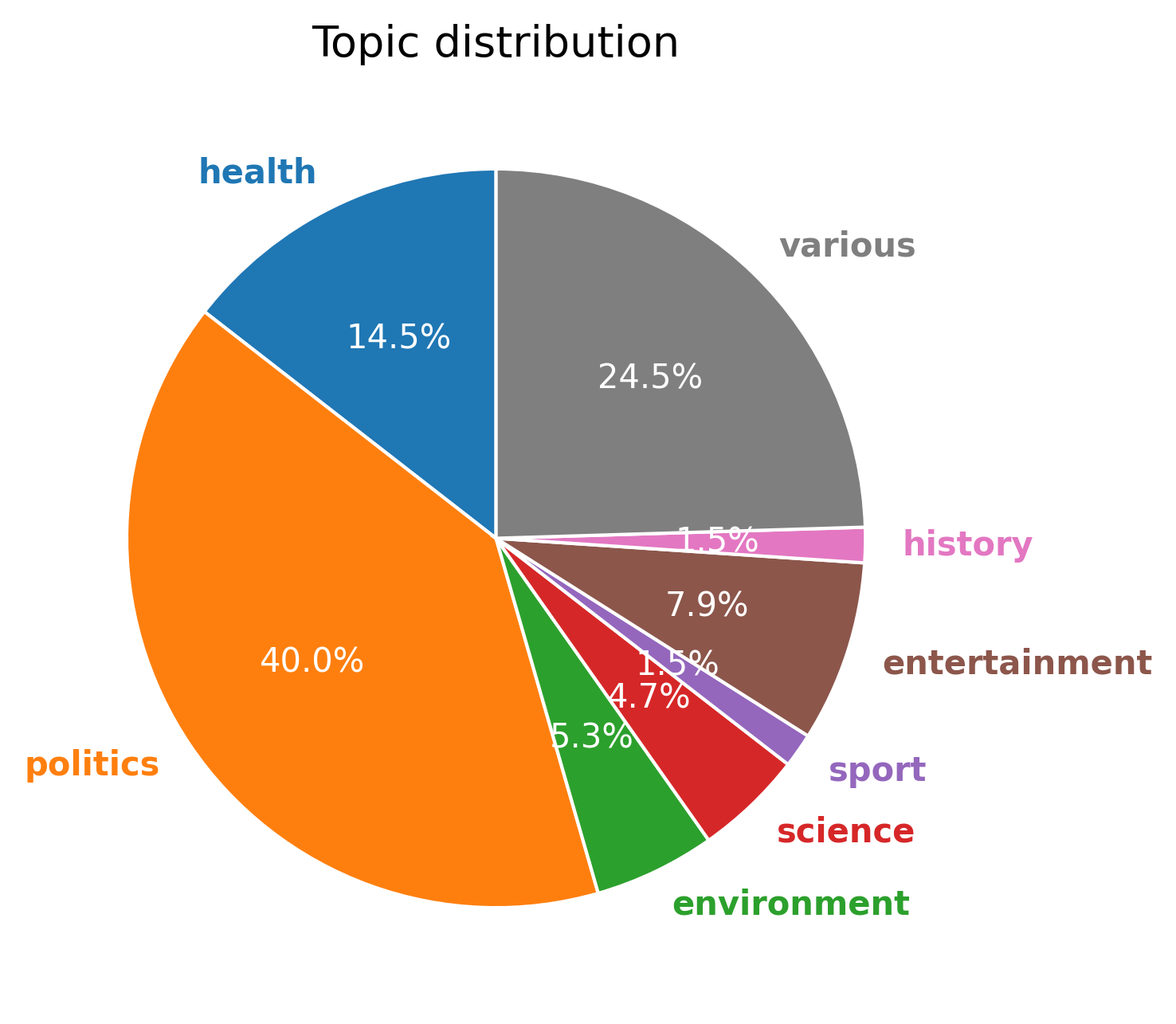}
    \caption{Topic distribution of the OOD dataset.}
    \label{fig:topic-distribution-ood}
\end{figure}

While strict topic balancing was not enforced, due to incomplete topic annotations in the source datasets, we conducted automatic topic detection over the final dataset using the Llama3:4B model. Topics were classified into seven predefined domains using the prompt described in Appendix~\ref{tab:topic_detection_prompt}. Figures~\ref{fig:topic-distribution-multicw} and~\ref{fig:topic-distribution-ood} present the resulting topic distributions. Consistent with the underlying source data, health (predominantly COVID-19–related) and political content constitute the majority of instances in both datasets, reflecting the domains most frequently targeted by professional fact-checking organizations. Importantly, the in-distribution and OOD datasets exhibit broadly comparable topic proportions, ensuring that out-of-distribution evaluation primarily probes linguistic generalization rather than domain shift. Detailed per-language topic statistics are provided in Appendix~\ref{sec:appendix-topic-detection}.

\begin{figure}[!ht]
    \centering
    \includegraphics[width=0.4\textwidth]{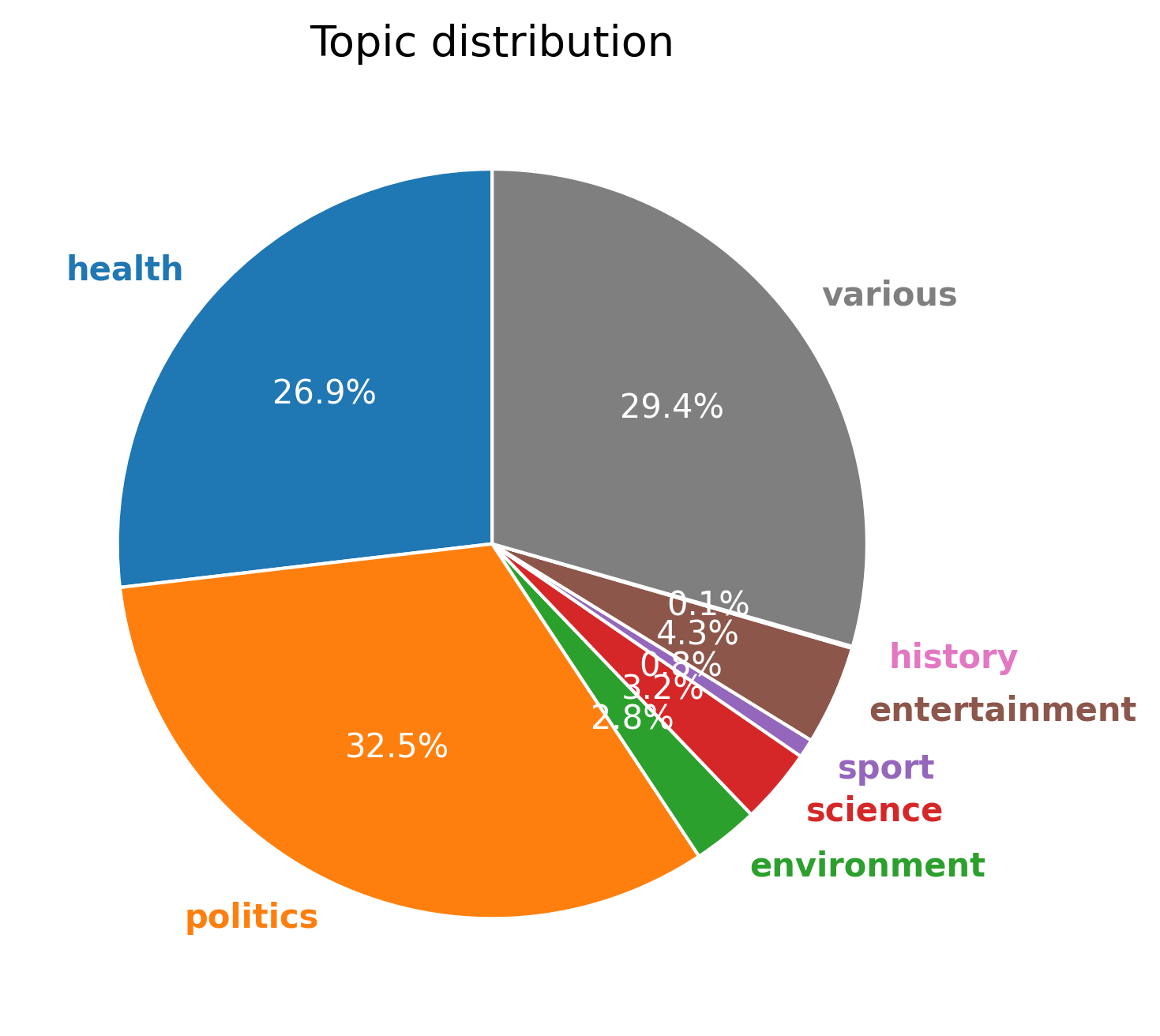}
    \caption{Topic distribution of the MultiCW dataset.}
    \label{fig:topic-distribution-multicw}
\end{figure}

\section{Experimental Setup}
\label{sec:experimental_setup}

\subsection{Fine-Tuned Transformer Models}
We use three multilingual Transformer models, each fine-tuned for five epochs on the MultiCW training set: 
\begin{itemize}[noitemsep, topsep=0pt]
    \item \textbf{XLM-R (base)} \cite{conneau2020unsupervised}, 
    a transformer pre-trained on 100 languages. 
    \item \textbf{mDeBERTa-v3 (base)} \cite{he2021deberta}, 
    a multilingual version of DeBERTa optimized with disentangled attention. 
    \item \textbf{LESA} \cite{gupta-etal-2021-lesa}, 
    a model based on the Linguistic Encapsulation and Semantic Amalgamation (LESA) architecture, 
    originally introduced for claim detection in noisy and heterogeneous text. 
\end{itemize}

\paragraph{Training Stability}
To ensure reproducibility and assess result stability, we trained each model three times with different random seeds. All other hyperparameters, data splits, and training conditions remained identical across runs. We report mean performance with standard deviations throughout our results. The training configuration of the fine-tuning process is described in detail in Appendix~\ref{sec:appendix-configuration}.

\subsection{Zero-shot LLMs}
\label{sec:methodology-prompts}
For comparison, we also evaluate a range of LLMs in a \textbf{zero-shot classification} setting, without task-specific fine-tuning. More specifically, we include 15 LLMs from 6 model families (Claude 3.x, GPT-4.1/GPT-o, Llama 3.x, Mistral, Qwen2.5, Nemotron-4). Each model received the same test inputs, framed via carefully designed prompts. We extracted binary decisions (\textit{claim} vs. \textit{non-claim}) from the models’ natural language responses 
using simple post-processing rules. If the response was ambiguous, a fallback clarification prompt was issued to enforce a strict Yes/No output.  

We designed and tested 6 prompt variants (see Appendix~\ref{sec:appendix-prompts} for more details), differing in reasoning style and explicitness of instructions, from which we selected two for in-depth evaluation. The prompts were developed through a multi-step process, integrating guidelines \cite{alam-etal-2021-fighting-covid}, LLM, and manual refinement. Initially, an LLM (specifically, Llama 3.1 405B) was used to generate a CoT (Chain-of-Thoughts) prompt based on the provided CLEF annotation guidelines. Subsequently, the prompt was manually adjusted to optimize its effectiveness. To select the most suitable \textbf{CoT prompt} from five alternatives, a small-scale pilot study was conducted, and the prompt that obtained the best results was chosen for further analysis. Additionally, the \textbf{GA (Guided Answer) prompt} was included in the comprehensive analysis, to investigate the capabilities of various LLMs when processing simple versus complex prompts, providing a more comprehensive understanding of their performance.

These prompt variants allow us to explore how different reasoning scaffolds influence LLM behavior. In all cases, models were constrained to return a minimal binary answer, ensuring comparability with supervised baselines.

\subsection{Evaluation Protocol}
We report results on the MultiCW test set using following metrics: \textbf{accuracy}, \textbf{precision macro} and \textbf{recall macro}.
Evaluations are conducted on: 
\begin{enumerate}[noitemsep, topsep=0pt]
    \item \textbf{Overall test set} (all samples).  
    \item \textbf{Noisy subset} (user-generated, less structured content).  
    \item \textbf{Structured subset} (well-formed text such as news articles).  
\end{enumerate}

This setup provides a clearer picture of model robustness by combining overall correctness (accuracy) with a fair evaluation across all classes (macro precision/recall), enabling us to assess whether fine-tuned models generalize across heterogeneous text sources as well as compare them with the performance of zero-shot LLMs.

\section{Results}
\label{sec:results}

We evaluate the performance of fine-tuned Transformer models (XLM-R, mDeBERTa, and LESA) on the MultiCW dataset and compare them against LLMs in a zero-shot setting. 

\subsection{Fine-Tuned Transformer Models}

Table~\ref{tab:transformer_multicw} presents the results for XLM-R, mDeBERTa, and LESA. All three models achieve competitive performance, although their relative strengths vary by writing style.

\begin{table}[ht]
\centering
\footnotesize
\setlength{\tabcolsep}{6pt}
\renewcommand{\arraystretch}{1.0}
\begin{tabular}{lccc}
\hline
\textbf{Metric} & \textbf{\shortstack{XLM-R}} & \textbf{mDeBERTa} & \textbf{LESA} \\
\hline
\multicolumn{4}{l}{\textbf{Overall}} \\
Accuracy          & 0.918 & \textbf{0.923} & 0.79 \\
Precision (macro) & 0.920 & \textbf{0.924} & 0.80 \\
Recall (macro)    & 0.918 & \textbf{0.923} & 0.79 \\
\hline
\multicolumn{4}{l}{\textbf{Noisy}} \\
Accuracy          & 0.872 & \textbf{0.875} & 0.70 \\
Precision (macro) & 0.875 & \textbf{0.877} & 0.72 \\
Recall (macro)    & 0.872 & \textbf{0.875} & 0.70 \\
\hline
\multicolumn{4}{l}{\textbf{Structured}} \\
Accuracy          & 0.966 & \textbf{0.972} & 0.88 \\
Precision (macro) & 0.967 & \textbf{0.973} & 0.88 \\
Recall (macro)    & 0.966 & \textbf{0.972} & 0.88 \\
\hline
\end{tabular}
\captionsetup[table]{skip=0pt}
\caption{Performance of fine-tuned Transformer models on the MultiCW test set (mean over 3 runs; all standard deviations $\leq 0.00$). Accuracy, macro precision, and macro recall are reported overall and by writing style.}
\vspace{-14pt}
\label{tab:transformer_multicw}
\end{table}

\paragraph{Observations.}  
XLM-R and mDeBERTa achieve strong overall accuracy (92\%), with mDeBERTa slightly ahead. LESA, despite being an earlier architecture, still generalizes across languages with 79\% accuracy, reflecting its effectiveness in combining semantic and syntactic features.

Performance differences are pronounced between writing styles. XLM-R and mDeBERTa exceed 97\% on structured claims but drop to 87--88\% on noisy ones. LESA follows the same trend but with a sharper decline: 88\% on structured versus only 70\% on noisy claims. This confirms that LESA struggles more with short, informal, or fragmented inputs, whereas modern Transformer backbones handle style variation more robustly.

Across all models, errors tend to be false negatives on noisy claims with implicit or sarcastic language, and false positives on structured claims reporting trivial factual statements. LESA, in particular, is more prone to misclassifying noisy text, highlighting the challenges of adapting linguistic-feature-based architectures to highly informal content.

To verify result stability, we trained each model three times with different random seeds. Standard deviations across runs were consistently low. On the noisy data of the dataset, XLM-R and mDeBERTa show standard deviations of $(\sigma = 0.0016)$ and $(\sigma = 0.00262)$, respectively, indicating highly stable training. On the structured data, standard deviations were  $(\sigma = 0.0016)$ and $(\sigma = 0)$, respectively. These small deviations confirm that performance differences between models are robust. Full per-language variance statistics are available in our repository.

\subsection{Out-of-Domain Evaluation}

To assess robustness beyond the original distribution, we evaluated all fine-tuned models on the MultiCW out-of-domain (OOD) set. Results are summarized in Table~\ref{tab:transformer_ood}.  

\begin{table}[ht]
\centering
\footnotesize
\setlength{\tabcolsep}{6pt}
\renewcommand{\arraystretch}{1.0}
\begin{tabular}{lccc}
\hline
\textbf{Metric} & \textbf{\shortstack{XLM-R}} & \textbf{mDeBERTa} & \textbf{LESA} \\
\hline
\multicolumn{3}{l}{\textbf{Overall}} \\
Accuracy          & 0.813 & \textbf{0.866} & 0.73 \\
Precision (macro) & 0.822 & \textbf{0.866} & 0.75 \\
Recall (macro)    & 0.827 & \textbf{0.875} & 0.75 \\
\hline
\multicolumn{3}{l}{\textbf{Noisy}} \\
Accuracy          & 0.734 & \textbf{0.830} & 0.65 \\
Precision (macro) & 0.759 & \textbf{0.836} & 0.69 \\
Recall (macro)    & 0.753 & \textbf{0.841} & 0.67 \\
\hline
\multicolumn{3}{l}{\textbf{Structured}} \\
Accuracy          & 0.895 & \textbf{0.902} & 0.82 \\
Precision (macro) & 0.892 & \textbf{0.899} & 0.82 \\
Recall (macro)    & 0.903 & \textbf{0.911} & 0.83 \\
\hline
\end{tabular}
\caption{Out-of-distribution evaluation of models fine-tuned on MultiCW. Accuracy, macro precision, and macro recall are reported overall and by writing style.}
\vspace{-10pt}
\label{tab:transformer_ood}
\end{table}

\paragraph{Observations.}  
Both XLM-R and mDeBERTa maintain accuracy of around 81-87\% in unseen languages and domains, confirming that pretrained multilingual Transformers adapt well when fine-tuned on diverse claim detection data. Their performance degradation relative to in-domain evaluation is modest (5--9 points).  

LESA, while still able to transfer, exhibits a more pronounced accuracy drop (to 73\%). Detailed metrics show recall remains relatively high (detecting most claims), but precision suffers, reflecting difficulties in filtering out non-claims in heterogeneous and stylistically diverse text. This is consistent with LESA’s reliance on manually crafted syntactic and semantic features, which are less stable under cross-domain shifts.

\subsection{Comparison with LLMs}

Beyond fine-tuned Transformers, we evaluated a range of open and commercial LLMs in a zero-shot setting. Table~\ref{tab:llm_results} summarizes their performance on the MultiCW test split containing both noisy and structured part.  

\begin{table}[ht]
\centering
\resizebox{\linewidth}{!}{
\begin{tabular}{lccc}
\hline
\makecell{\textbf{Model}} &
\makecell{\textbf{Accuracy}} &
\makecell{\textbf{Precision} \\ \textbf{macro}} &
\makecell{\textbf{Recall} \\ \textbf{macro}} \\
\hline
Claude 3.5 Haiku (CoT)          & \textbf{0.75} & \textbf{0.76} & \textbf{0.75} \\
Claude 3.7 Sonnet (CoT)         & 0.72 & 0.76 & 0.72 \\
GPT-4.1 (CoT)                   & 0.65 & 0.71 & 0.65 \\
GPT-4.1 Mini (CoT)              & 0.65 & 0.69 & 0.65 \\
GPT-4o (CoT)                    & 0.69 & 0.72 & 0.69 \\
GPT-4o Mini (CoT)               & 0.69 & 0.71 & 0.69 \\
Llama 3.1 70B (CoT)             & \textbf{0.74} & \textbf{0.74} & \textbf{0.74} \\
Llama 3.1 8B (CoT)              & 0.69 & 0.71 & 0.69 \\
Llama 3.2 1B (CoT)              & 0.53 & 0.53 & 0.53 \\
Llama 3.2 3B (CoT)              & 0.52 & 0.53 & 0.52 \\
Llama 3.3 70B (CoT)             & 0.71 & 0.75 & 0.71 \\
Mistral Large 123B (CoT)        & 0.69 & 0.70 & 0.69 \\
Mistral Nemo 12B (CoT)          & 0.66 & 0.66 & 0.66 \\
Qwen2.5 72B (CoT)               & 0.70 & 0.73 & 0.70 \\
Nemotron-4 340B (CoT)           & \textbf{0.79} & \textbf{0.80} & \textbf{0.79} \\
Claude 3.5 Haiku (GA)           & 0.59 & 0.59 & 0.59 \\
Claude 3.7 Sonnet (GA)          & 0.61 & 0.64 & 0.61 \\
GPT-4.1 (GA)                    & 0.73 & 0.73 & 0.73 \\
GPT-4.1 Mini (GA)               & 0.68 & 0.68 & 0.68 \\
GPT-4o (GA)                     & 0.72 & 0.75 & 0.72 \\
GPT-4o Mini (GA)                & 0.76 & 0.76 & 0.76 \\
Llama 3.1 70B (GA)              & 0.65 & 0.65 & 0.65 \\
Llama 3.1 8B (GA)               & 0.62 & 0.63 & 0.62 \\
Llama 3.2 1B (GA)               & 0.48 & 0.45 & 0.48 \\
Llama 3.2 3B (GA)               & 0.52 & 0.62 & 0.52 \\
Llama 3.3 70B (GA)              & 0.67 & 0.69 & 0.67 \\
Mistral Large 123B (GA)         & 0.64 & 0.67 & 0.64 \\
Mistral Nemo 12B (GA)           & 0.65 & 0.65 & 0.65 \\
Qwen2.5 72B (GA)                & 0.69 & 0.70 & 0.69 \\
Nemotron-4 340B  (GA)           & 0.72 & 0.73 & 0.72 \\
\hline
\end{tabular}}
\caption{Zero-shot performance of large language models on the MultiCW test set. The top 3 best results are highlighted in bold.}
\label{tab:llm_results}
\vspace{-10pt}
\end{table}

\paragraph{Observations.}  
The results show that the best-performing LLM is \textbf{Nemotron-4 340B (CoT)}, achieving the highest accuracy at 79\%, followed by \textbf{Claude 3.5 Haiku (CoT)} at 75\% and \textbf{Llama 3.1 70B (CoT)} at 74\%. Other strong contenders include Claude 3.7 Sonnet (CoT) at 72\% and GPT-4.1 (GA) at 73\%. Interestingly, CoT prompting shows inconsistent effects across models: while it significantly improves performance for some models (e.g., Claude 3.5 Haiku from 59\% (GA) to 75\% (CoT)), others perform better without it (e.g., GPT-4.1 drops from 73\% to 65\% with CoT). In contrast, smaller models such as Llama 3.2 1B and 3B perform poorly, with accuracy levels around 48--53\%. In the table ~\ref{tab:llm_results}, ``CoT'' refers to the \textit{CoT\_CLEF\_on\_Q} prompt, which was selected as the best-performing prompt.

Overall, fine-tuned Transformers still outperform LLMs on in-domain test data (92\% for XLM-R and mDeBERTa, 79\% for LESA, see Table~\ref{tab:transformer_multicw}). However, the strongest LLMs are now able to close much of the gap, especially on structured claims, with Nemotron (CoT) and Claude 3.5 Haiku (CoT) showing competitive performance. This suggests that hybrid strategies (e.g., few-shot prompting combined with style normalization) could further boost zero-shot performance and reduce reliance on fine-tuning. This is also in line with results on other tasks presented in related works, e.g., in~\cite{pecher2025comparingspecialisedsmallgeneral}.

\subsection{Per-Language Performance}

Detailed per-language results for all fine-tuned models and the top-performing LLMs are provided in Appendix~\ref{sec:appendix-per-language-evaluation}. We observe substantial variation across languages, with accuracy differences exceeding 20 percentage points in some cases.

\paragraph{Observations.}
Fine-tuned transformers perform best on Western European languages (Portuguese, French, German: 96--98\% accuracy) and struggle most with Arabic (80--81\%) and slightly with Slavic and Central European languages (90--95\%). This pattern reflects structural differences in how claims are expressed across languages.

LLMs show a different distribution of strengths: while they also excel on major European languages, their performance on languages like Bengali and Hindi is more competitive with fine-tuned models than expected, suggesting better cross-lingual transfer in zero-shot settings. However, LLMs show high variability, but with all LLMs struggling with Bulgarian and Arabic.

These language-specific patterns highlight the importance of evaluating multilingual claim detection systems across diverse linguistic typologies, not just high-resource languages. Full results are presented in Tables~\ref{tab:transformer_per_language} and~\ref{tab:llms_per_language}.

\section{Discussion}
\label{sec:discussion}

\paragraph{Value of the MultiCW Dataset.}
The construction of the MultiCW dataset proved central to the experimental analysis. 
By balancing languages, classes, and writing styles, we ensured that models were not biased 
towards particular conditions. The inclusion of both \textit{noisy} (social-media like) and 
\textit{structured} (encyclopedic and news-like) subsets enabled systematic evaluation of 
cross-style robustness. Results consistently confirmed that noisy text is the most challenging 
condition, highlighting the importance of datasets that reflect the heterogeneous nature 
of real-world claims.

\paragraph{Performance of Fine-Tuned Transformers.}
Fine-tuned multilingual Transformers achieved the strongest results overall. 
Both \textbf{XLM-R} and \textbf{mDeBERTa} reached accuracies close to 92\%, 
with balanced precision and recall across classes. 
Their robustness across languages and styles demonstrates the benefit of large-scale multilingual pretraining.  
\textbf{LESA}, while conceptually appealing due to its integration of syntactic and semantic 
representations, underperformed compared to modern Transformer baselines, particularly on 
noisy text. 

\paragraph{Comparison with LLMs.}
Large language models evaluated in zero-shot settings achieved solid but consistently 
lower performance than fine-tuned models. The best-performing LLMs, such as 
\textbf{Nemotron-4 340B (CoT)} and \textbf{Claude 3.5 Haiku (CoT)}, reached accuracies 
around 75--79\%, trailing behind the fine-tuned Transformers by 13--17 points. 
Smaller LLMs (e.g., Llama 3.2 1B/3B) performed significantly worse, often dropping below 55\% accuracy.  

These results highlight two key observations: 
\begin{enumerate}[noitemsep, topsep=0pt]
    \item \textbf{Task-specific fine-tuning remains crucial}: Even relatively small Transformers, 
    once adapted to the dataset, outperform powerful LLMs operating in a zero-shot setting. 
    \item \textbf{Prompting format}: Among the six prompt designs, structured Chain-of-Thought (CoT) prompts (especially \texttt{CoT\_CLEF\_N}) achieved the highest accuracy, outperforming both minimal prompts and alternative designs. This shows that explicitly guiding LLMs through intermediate reasoning steps is particularly effective for CW detection, where alignment with evidence-based criteria is critical.
\end{enumerate}

\paragraph{Generalization to New Languages.}
Evaluation on an extended multilingual set confirmed that fine-tuned models, especially 
XLM-R and mDeBERTa, generalize well to unseen languages.
This underscores the advantage of Transformer-based pretraining on large multilingual corpora, while also suggesting that 
linguistically informed models may require additional adaptation to maintain performance across highly diverse settings.


\section{Conclusion}
\label{sec:conclusion}

In this work, we introduced the \textbf{MultiCW dataset}, a large-scale multilingual resource for claim detection, balanced across languages, classes, and writing styles and we provided baseline results of the most common Transformer models. By incorporating both \textit{noisy} and \textit{structured} subsets, the dataset enabled a systematic investigation of model robustness under heterogeneous conditions.

Our experiments demonstrated that \textbf{fine-tuned multilingual Transformers}, specifically XLM-R and mDeBERTa, achieve the strongest performance, reaching accuracies of 92\% across the test set. \textbf{LESA}, a linguistically informed architecture, performed competitively on structured inputs but lagged behind on noisy data.

In contrast, \textbf{large language models} evaluated in a zero-shot setting underperformed relative to fine-tuned Transformers, with best-case accuracies around 75-79\%. Nevertheless, carefully designed CoT prompts improved their effectiveness, suggesting that prompt engineering and utilization of in-context learning or instruction tuning (with PEFTs like LoRA) can partially bridge the gap. Extended multilingual evaluation confirmed the stability of fine-tuned Transformers, with only minor accuracy drops on unseen languages.

Overall, our findings highlight that task-specific fine-tuning remains crucial for robust claim detection, and LLMs, while not yet competitive in zero-shot claim detection, hold promise to improve in the future through more advanced prompting strategies and targeted fine-tuning.

Future research should explore hybrid systems that combine the precision of fine-tuned models with the adaptability of LLMs, as well as investigate more advanced multilingual alignment and domain adaptation strategies to further enhance robustness in real-world fact-checking applications.

\section*{Acknowledgement}
This work was partially supported by the European Union under the Horizon Europe projects: \textit{vera.ai} (GA No. \href{https://doi.org/10.3030/101070093}{101070093}), and by \textit{AI-CODE} (GA No. \href{https://cordis.europa.eu/project/id/101135437}{101135437}); and by EU NextGenerationEU through the Recovery and Resilience Plan for Slovakia under the project No. 09I01-03-V04-00006.

\section*{Limitations}
\label{sec:limitations}

The MultiCW dataset and benchmarks establish a solid foundation for research in multilingual check-worthiness detection, yet several limitations remain.

First, although MultiCW covers 16 languages, seven domains, and two styles, it is not exhaustive---many African/Indigenous languages and domains such as finance or law remain underrepresented. For low‑resource languages we used machine translation to fill the gaps; this can introduce subtle biases compared with naturally authored text. After creation we performed several checks---native‑speaker spot‑checks, automated bias tests, and consistency audits---to confirm that any residual bias is minimal relative to the broader coverage achieved.

Second, our definition of check-worthiness follows a binary framing (\textit{check-worthy} vs. \textit{non-check-worthy}). Our dataset does not yet capture a graded notion, which could provide richer supervision for real-world fact-checking systems.

Third, while we balance between \textit{noisy} and \textit{structured} styles, other forms of variation remain unexplored. For example, code-switching, dialectal variation, and multimodal inputs (e.g., text paired with images) are increasingly common in misinformation but are not represented in MultiCW. This limits the generalization of models trained solely on our dataset.

Finally, our experiments focus on Transformer-based models (XLM-R, mDeBERTa, LESA) and a set of LLMs in zero-shot settings. We do not explore fine-tuning of LLMs, few-shot prompting, or hybrid retrieval-based approaches, all of which could further close the gap between supervised and zero-shot performance.

\section*{Ethical Considerations}
\label{sec:ethics}
We have performed a thorough ethical assessment of all aspects of our work (data, processes, model), using an extended ethics checklist. Regarding the created MultiCW dataset published with this work, we partially republish already existing datasets. However, we use them in accordance with their intended purposes and licenses. Also, the republished portions maintain the same access restrictions as the original datasets through our Zenodo repository. Additionally, our GitHub repository provides the tools and mechanisms to process the original datasets into the format used in our work, enabling replication of our results.

Regarding the development of robust check-worthy claim detection models, there is a potential concern that the AI models, if not transparently communicated, may generate uncertainty about their capabilities, leading to over-reliance on their use. Establishing mechanisms for auditability, technical robustness, and safety poses a significant challenge but is crucial to prevent unintended consequences.


\bibliography{custom}

\appendix

\section{Source Datasets}
\label{sec:appendix-datasets}

The MultiCW dataset integrates and balances samples from multiple publicly available corpora 
to ensure diversity in language, topic, and writing style. 
We summarize each source dataset and its role in MultiCW below.  

\subsection{CLEF-2022 CheckThat! Lab}
\begin{figure}[!ht]
    \centering
    \includegraphics[width=0.4\textwidth]{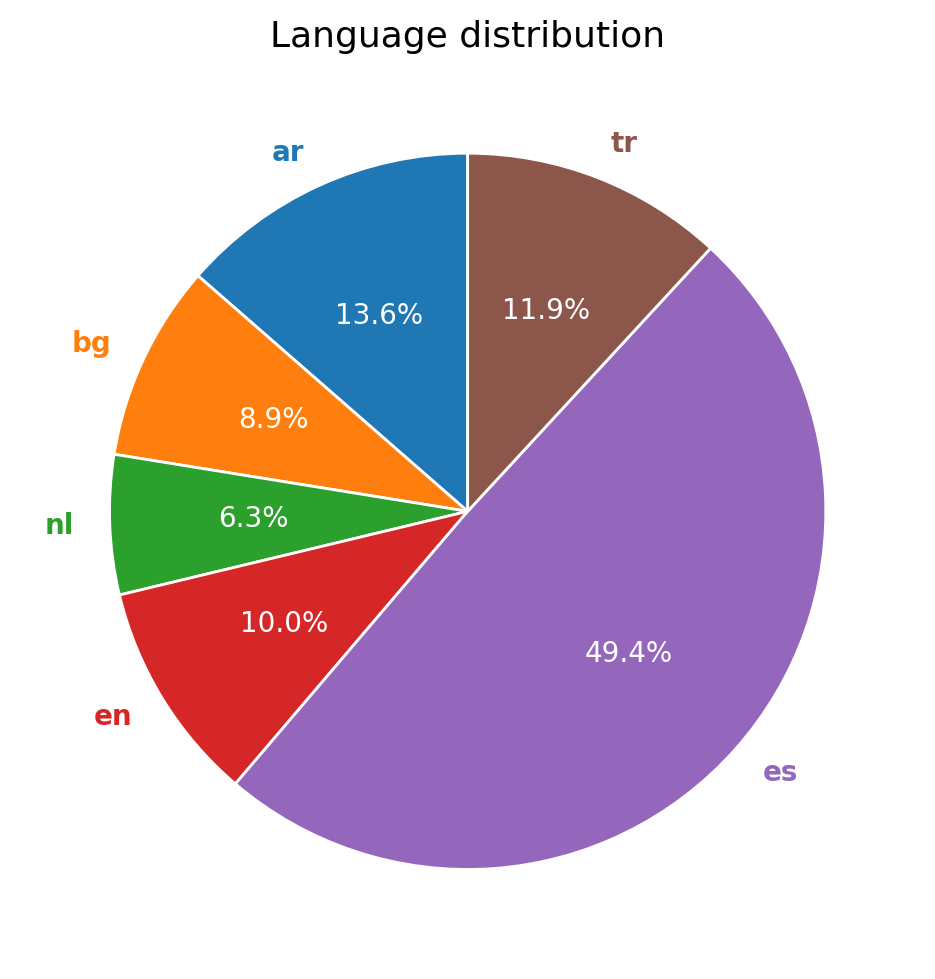}
    \caption{Language distribution of the CLEF-2022 dataset.}
    \label{fig:lang_distribution_clef2022}
\end{figure}

\textbf{CLEF-2022} \cite{DBLP:conf/clef/NakovBMAMCKZLSM22} provides multilingual datasets 
for check-worthiness detection, covering six languages (Arabic, Bulgarian, Dutch, English, Spanish, Turkish). 
It contains 30k claims annotated as check-worthy or not, mainly drawn from social media and news. 
The dataset is valuable for its multilingual scope, but exhibits class imbalance (e.g., Spanish dominates the distribution, see Figure~\ref{fig:lang_distribution_clef2022}).

\subsection{CLEF-2023 CheckThat! Lab}
\begin{figure}[!ht]
    \centering
    \includegraphics[width=0.4\textwidth]{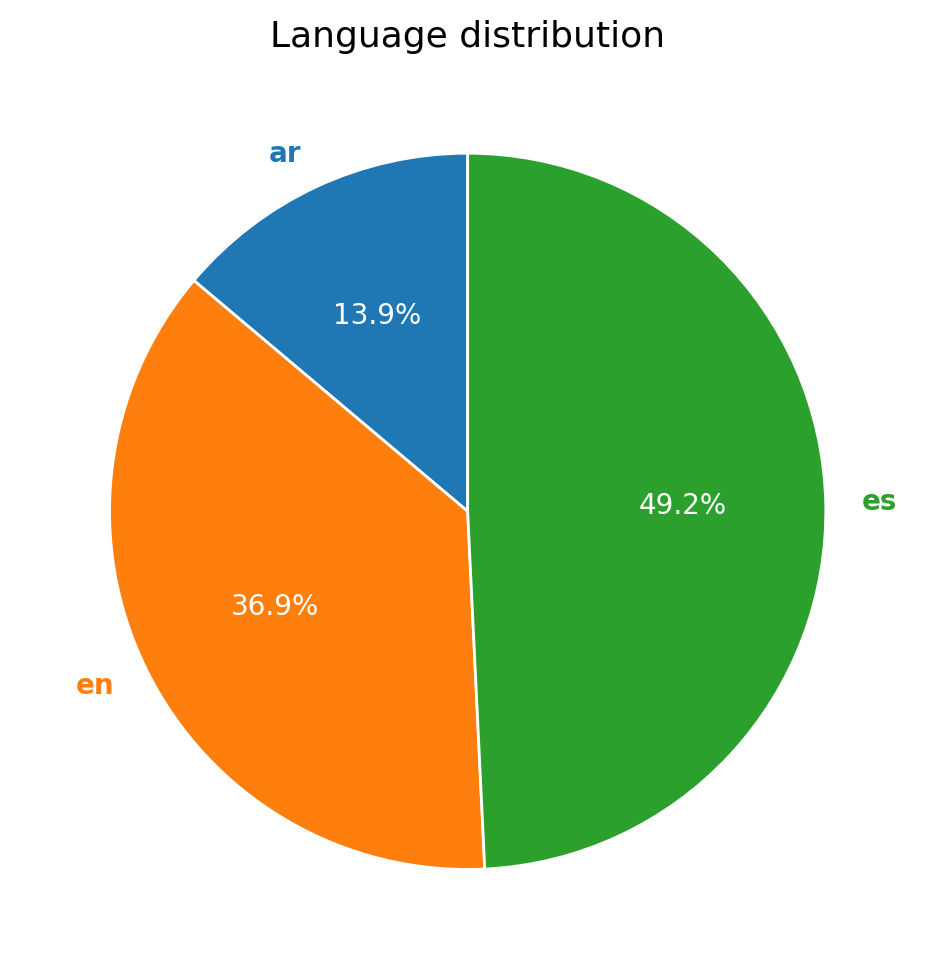}
    \caption{Language distribution of the CLEF-2023 dataset.}
    \label{fig:lang_distribution_clef2023}
\end{figure}

\textbf{CLEF-2023} \cite{clef-checkthat:2023} extended the task to additional languages 
and improved the balance of positive and negative samples. Figure~\ref{fig:lang_distribution_clef2023} shows its language distribution. 
It retains the same structure as CLEF-2022 but incorporates broader topical coverage 
across politics, health, and social issues.  

\subsection{MultiClaim}
The \textbf{MultiClaim dataset} \cite{pikuliak-etal-2023-multilingual} is a multilingual benchmark for claim detection, comprising both fact-checking articles (structured style) and social media posts (noisy style).
In our work, it was used in two configurations:  
(1) \textbf{zenodo} — the original MultiClaim dataset released on Zenodo, and  
(2) \textbf{extended} — an expanded version we constructed using several of the original data sources employed in MultiClaim.  

\begin{figure}[!ht]
    \centering
    \includegraphics[width=0.4\textwidth]{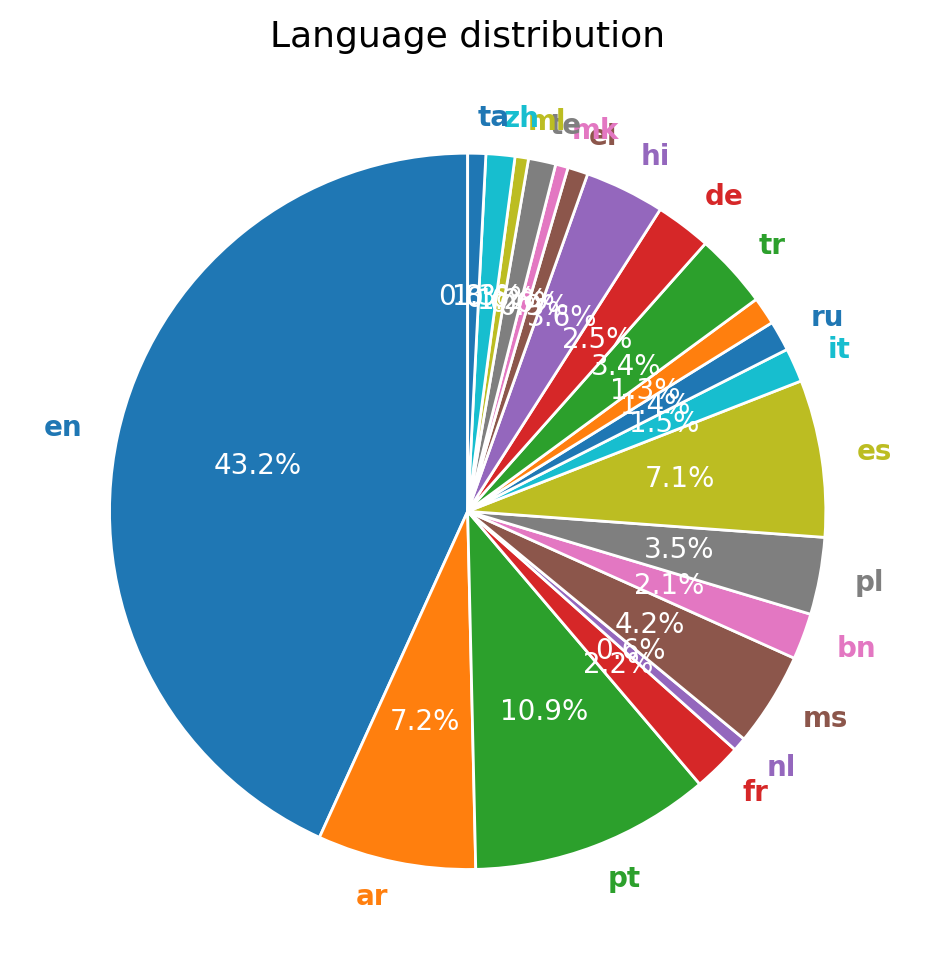}
    \caption{Language distribution of the original MultiClaim dataset.}
    \label{fig:lang_distribution_multiclaim}
\end{figure}

\begin{figure}[!ht]
    \centering
    \includegraphics[width=0.4\textwidth]{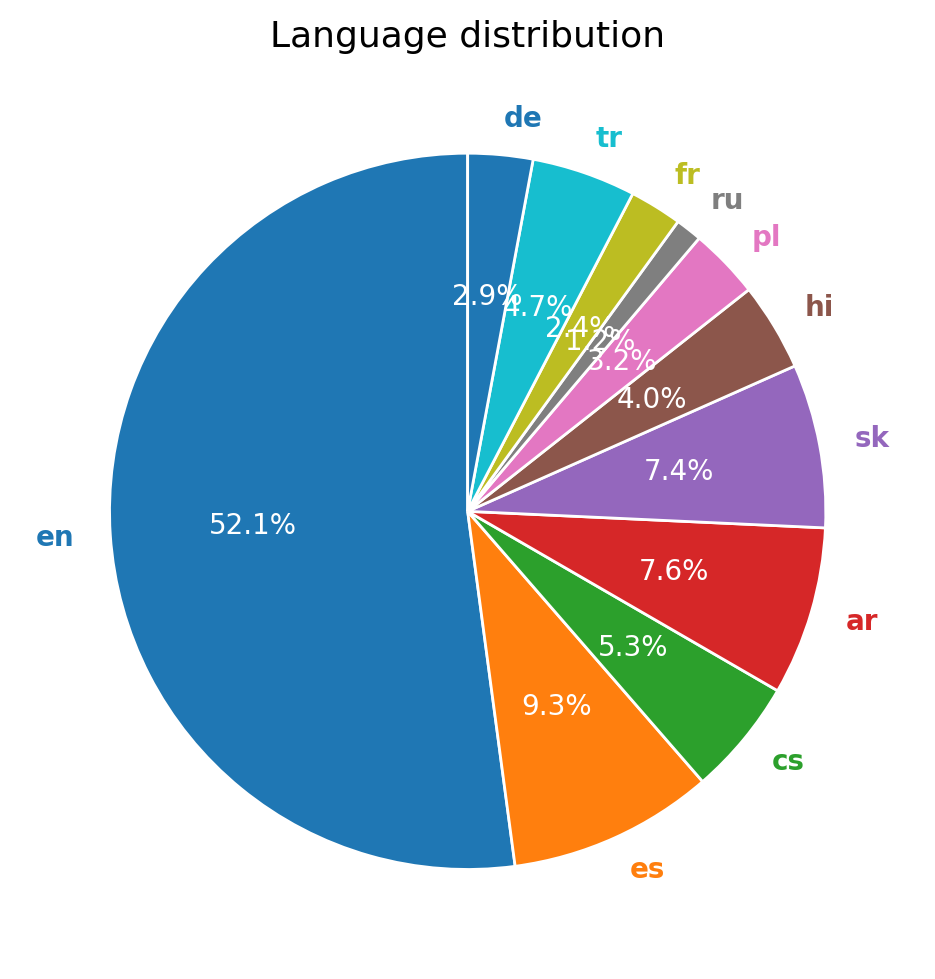}
    \caption{Language distribution of the extended MultiClaim dataset.}
    \label{fig:lang_distribution_multiclaim_extended}
\end{figure}

Together, these configurations provide broad multilingual coverage (see Figures~\ref{fig:lang_distribution_multiclaim} and~\ref{fig:lang_distribution_multiclaim_extended}), featuring fact-checker–curated positive instances while requiring additional balancing to mitigate language skew.  
Notably, after our experiments, a subsequent version of MultiClaim was published \cite{moro_2025_multiclaim2} that also incorporated our extended data.

\subsection{Ru22Facts}
\begin{figure}[!ht]
    \centering
    \includegraphics[width=0.4\textwidth]{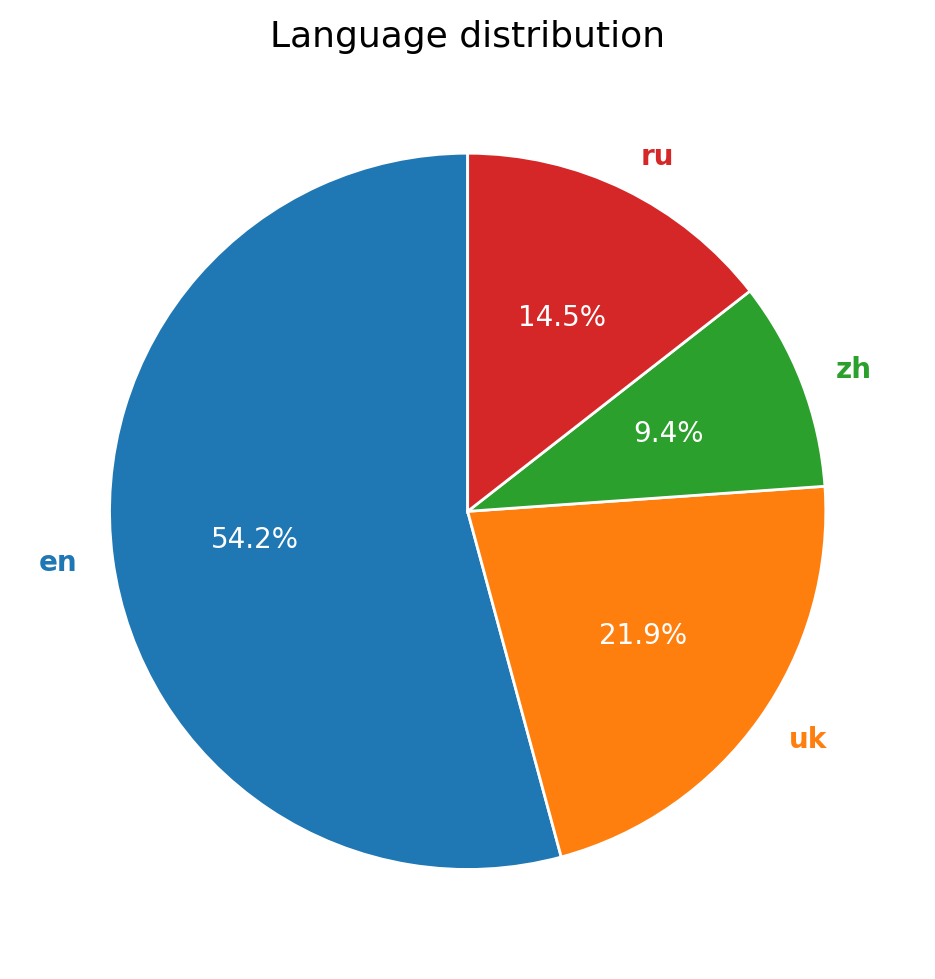}
    \caption{Language distribution of the Ru22Fact dataset.}
    \label{fig:lang_distribution_ru22fact}
\end{figure}

\textbf{Ru22Facts dataset} \cite{zeng-etal-2024-ru22fact} constitutes a multilingual fact-checking dataset on the Russia-Ukraine conflict in 2022 of 16K samples, each containing real-world claims, optimized evidence, and referenced explanation. Figure~\ref{fig:lang_distribution_ru22fact} shows its language distribution. The authors also developed an end-to-end explainable fact-checking system to verify claims and generate explanations.

\subsection{LESA (EACL-2021)}
The \textbf{LESA dataset} \cite{gupta-etal-2021-lesa} contains $\sim$40k English samples 
from multiple domains and styles: Twitter (noisy), LiveJournal and Wiki Talk Pages (semi-structured), 
and Persuasive Essays, Web Discourse, and other sources (structured).  
Because some noisy/semi samples did not meet our stricter check-worthiness definition, 
we filtered them using an LLM-based validation step, retaining $\sim$27k balanced examples.  
LESA served as both a direct source of English claims and a basis for machine translations into low-resource languages.  

\begin{figure}[ht]
    \centering
    \includegraphics[width=0.4\textwidth]{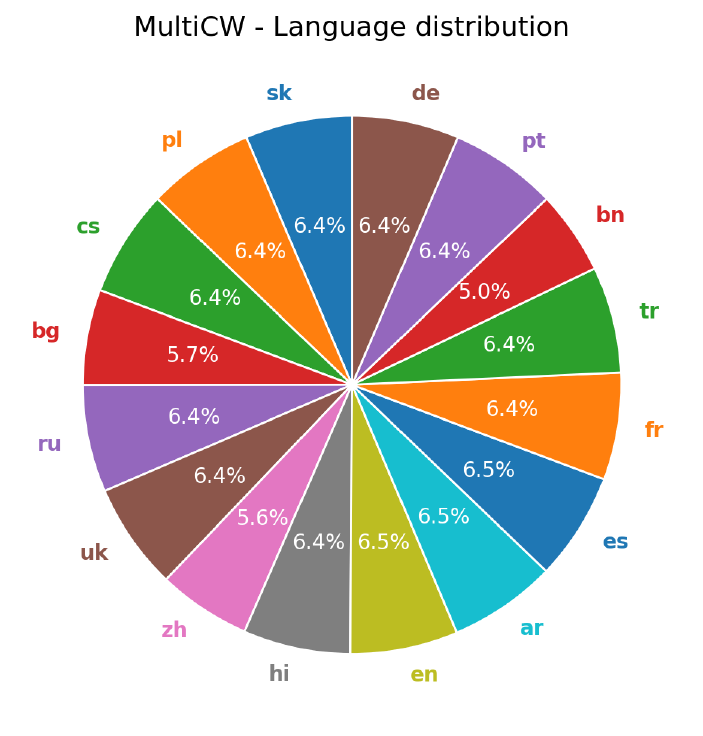}
    \caption{Balanced MultiCW dataset language distribution.}
    \label{fig:lang_distribution_multicw}
\end{figure}

\subsection{Wikipedia Samples}
To supply neutral and non-check-worthy structured claims, we scraped Wikipedia across 16 languages.  
Named entities were first extracted (using GLiNER \cite{zaratiana2023gliner}), 
then entity-relevant sentences were retrieved and proportionally sampled 
to reflect topical frequency distributions observed in other datasets.  
These samples enhance topic diversity and reduce class imbalance in structured text.  

\subsection{Data Provenance Labels}
Every MultiCW sample is tagged with a provenance label indicating its origin:  
\begin{itemize}[noitemsep, topsep=0pt]
    \item \textbf{Manual}: curated from fact-checking datasets (e.g., CLEF, Ru22Facts).  
    \item \textbf{Augmented}: translated from English datasets (LESA, ClaimBuster).  
    \item \textbf{Wikipedia}: extracted from encyclopedic text using NER-driven scraping.  
\end{itemize}
This labeling ensures transparency and allows researchers to analyze results across different data sources.

\section{Detailed prompts for LLM Experiments}
\label{sec:appendix-prompts}
Here we provide a detailed description of all the prompts used in LLM evaluation process. We designed and tested six prompt variants that we obtained by multi-step process described in Section \ref{sec:methodology-prompts}:

\begin{itemize}[noitemsep, topsep=0pt]
    \item \textbf{GA (Guided Answer)}: A concise prompt instructing the model to act as a fact-checker 
    and classify sentences strictly as \texttt{Yes} (check-worthy) or \texttt{No} (not check-worthy). 
    See Table~\ref{tab:prompt-ga}.
    
    \item \textbf{CoT\_CLEF\_on\_Q}: A step-by-step Chain-of-Thought (CoT) style prompt 
    inspired by CLEF fact-checking guidelines \cite{alam-etal-2021-fighting-covid}, asking the model to evaluate claims along dimensions 
    of verifiability, harm, public interest, and evidence before producing a binary judgment. 
    See Table~\ref{tab:prompt-cot-clef-on-q}.
    
    \item \textbf{CoT\_H\_on\_Q}: A hierarchical CoT prompt focusing on implications for policy, safety, 
    health, or public debate, with explicit intermediate Yes/No questions leading to a final binary decision. 
    See Table~\ref{tab:prompt-cot-h-on-q}.
    
    \item \textbf{CoT\_CLEF\_N}: A refined CoT prompt with multi-scale evaluation (likelihood of falsehood, 
    public interest, potential harm, all rated on a 1–5 scale) followed by a final binary decision 
    (\texttt{claim} vs. \texttt{non-claim}). 
    See Table~\ref{tab:prompt-cot-clef-n}.
    
    \item \textbf{HCoT}: A hybrid prompt combining guided sub-questions and reasoning steps 
    (e.g., verifiability, opinion vs. fact, sensitivity of topic, harmfulness) with a final Yes/No classification. 
    See Table~\ref{tab:prompt-hcot}.
    
    \item \textbf{CoT\_CLEF\_2N}: A variant of the CLEF-inspired CoT, emphasizing multilingual robustness. 
    It explicitly instructs the model to consider claims across 16 languages and judge check-worthiness 
    based on falsehood likelihood, public interest, and harmfulness, before issuing a binary label. 
    See Table~\ref{tab:prompt-cot-clef-2n}.
\end{itemize}

\begin{table}[ht!]
\centering
\small
\begin{tabular}{|p{0.45\textwidth}|}
\hline
\textbf{System message:} You are a fact-checker assistant with a task to identify sentences that are check-worthy. Sentence is check-worthy only if it contains a verifiable factual claim and that claim can be harmful. \\
\hline
\textbf{User message:} Classify the check-worthiness of these sentences outputting only Yes or No: [sentence] do not display any explanations \\
\hline
\end{tabular}
\caption{GA (Guided Answer) prompt template.}
\label{tab:prompt-ga}
\vspace{-10pt}
\end{table}

\begin{table}[ht!]
\centering
\small
\begin{tabular}{|p{0.45\textwidth}|}
\hline
\textbf{System message:} You will be presented with a text and asked to determine if it contains a check-worthy claim. To make this determination, you will follow a series of steps and answer a set of questions. Your final answer will be "Yes" if the text contains a check-worthy claim and "No" if it does not. \\
\hline
\textbf{User message:} Text: [sentence]

Chain of Thoughts (CoT) Steps:

Step 1: Read the text carefully and identify any claims or statements that could be verified or debunked by a fact-checker.

Step 2: Consider the likelihood that the claim in the text is false or misleading. Ask yourself: Is the claim suspicious or too good (or bad) to be true?

Step 3: Evaluate the significance of the claim. Is it a matter of public interest or a trivial/personal matter? Would the general public be interested in knowing whether the claim is true or false?

Step 4: Assess the potential impact of the claim if it is false or misleading. Could it harm individuals, organizations, or society as a whole?

Step 5: Determine if the text appears to be spreading rumours or misinformation about a particular topic or individual.

Step 6: Check if the claim is supported by credible sources or evidence. Are there any references or citations provided to back up the claim?

Step 7: Analyse the tone and language used in the text. Is it emotive or sensationalist? Could it be intended to manipulate or deceive readers?

Step 8: Consider the topic of the claim. Is it related to a matter of public concern, such as healthcare, politics, or current events?

Step 9: Evaluate the potential harm that the claim could cause if it is false or misleading. Could it be used to discredit individuals, organizations, or products?

Step 10: Finally, assess whether the text provides enough context and information for a reader to make an informed decision about the claim's validity.

Final Step: Based on your answers to the previous steps, determine if the text contains a check-worthy claim. If you answered "yes" to any of the following questions - 1, 4, 5, 7, 8, or 9 - or if you answered "no" to questions 6 or 10, then the text likely contains a check-worthy claim. Otherwise, the text does not contain a check-worthy claim.

Response Format: Please respond with a simple "Yes" or "No" to indicate whether the text contains a check-worthy claim. Do not display any explanations \\
\hline
\end{tabular}
\caption{CoT\_CLEF\_on\_Q prompt template.}
\label{tab:prompt-cot-clef-on-q}
\vspace{-15pt}
\end{table}

\begin{table}[ht!]
\centering
\small
\begin{tabular}{|p{0.45\textwidth}|}
\hline
\textbf{System message:} You will be given a text and asked to determine if it contains a check-worthy claim. A check-worthy claim is a statement that has significant implications for public policy, public safety, public health, or societal well-being, or is controversial, contentious, disputed, or likely to spark debate among experts or the public. To make this determination, you will follow a series of steps and answer a set of questions. Please answer each question with either "Yes" or "No". \\
\hline
\textbf{User message:} Text: [sentence]

Chain of Thoughts (CoT) Steps:

Step 1: Does the claim have significant implications for public policy, public safety, public health, or societal well-being? Please answer with either "Yes" or "No".

Step 2: Is the claim controversial, contentious, disputed, or likely to spark debate among experts or the public? Please answer with either "Yes" or "No".

Step 3: Does the claim have the potential to impact a lot of people, influence public opinion, shape decisions, or drive behavior? Please answer with either "Yes" or "No".

Step 4: Is the source of this claim a public figure, authoritative source, or institution with influence? Please answer with either "Yes" or "No".

Step 5: Is the claim purely subjective, such as a matter of personal opinion? Please answer with either "Yes" or "No".

Step 6: Is the claim about trivial or inconsequential matters that do not affect broader understanding or decision-making of many people? Please answer with either "Yes" or "No".

Step 7: Does the claim consist of common knowledge and align with widely accepted facts or common knowledge, and does not introduce new information or controversy? Please answer with either "Yes" or "No".

Step 8: Is the claim about some insignificant named entity or is it related to some insignificant named entity? Please answer with either "Yes" or "No".

Step 9: Does the claim lack public significance, lack impact on a lot of people, or lack potential harm for society? Please answer with either "Yes" or "No".

Step 10: Based on your answers to the previous questions, determine if the text contains a check-worthy claim. If you answered "Yes" to any of questions 1-4, and "No" to all of questions 5-9, then the text contains a check-worthy claim, and the answer is "Yes". If you answered "No" to all of questions 1-4, or "Yes" to any of questions 5-9, then the text does not contain a check-worthy claim, and the answer is "No". Please respond with either "Yes" or "No". Do not display any explanations \\
\hline
\end{tabular}
\caption{CoT\_H\_on\_Q prompt template.}
\label{tab:prompt-cot-h-on-q}
\vspace{-20pt}
\end{table}

\begin{table}[ht!]
\centering
\small
\begin{tabular}{|p{0.45\textwidth}|}
\hline
\textbf{System message:} Determine whether a sentence contains a claim that should be verified by a professional fact-checker. To make this determination, follow the Chain of Thoughts (CoT) approach:

Chain of Thoughts (CoT) Approach:
1. Read the sentence carefully and analyze its content.
2. Determine if the sentence contains a claim that is likely to be false, of public interest, and/or appears to be harmful.
3. Assess the likelihood that the sentence contains false information (Scale: 1-5, where 1 is "NO, definitely contains no false information" and 5 is "YES, definitely contains false information").
4. Evaluate the impact of the sentence's claim on the general public (Scale: 1-5, where 1 is "NO, definitely not of interest" and 5 is "YES, definitely of interest").
5. Consider the potential harm caused by the sentence to society, person(s), company(s), or product(s) (Scale: 1-5, where 1 is "NO, definitely not harmful" and 5 is "YES, definitely harmful").
6. If the sentence contains a claim that is likely to be false, of public interest, and/or appears to be harmful, then it is check-worthy.
7. If the sentence is check-worthy, determine if a professional fact-checker should verify the claim.
8. Consider the potential consequences of not verifying the claim.
9. Evaluate the potential benefits of verifying the claim.
10. Based on the analysis, determine whether the sentence contains a claim that should be verified by a professional fact-checker.

Final Answer: Based on the CoT approach, determine whether the sentence contains a claim that should be verified by a professional fact-checker. Your final answer should be either "claim" or "non-claim". \\
\hline
\textbf{User message:} Please analyze the following sentence using the Chain of Thoughts (CoT) approach and determine whether it contains a claim that should be verified by a professional fact-checker.

Sentences: [sentence]

[CoT steps repeated]

Our final answer is, do not display any explanations or intermediate steps: \\
\hline
\end{tabular}
\caption{CoT\_CLEF\_N prompt template.}
\label{tab:prompt-cot-clef-n}
\vspace{-20pt}
\end{table}

\begin{table}[ht]
\centering
\small
\begin{tabular}{|p{0.45\textwidth}|}
\hline
\textbf{Prompt:} Question: Classify if the sentence is check-worthy: They have a VAT tax.

Are follow up questions needed here: Yes

Follow up: Does the sentence contain a verifiable factual claim?
Intermediate answer: Yes

Follow up: Could the claim be considered a statement of opinion?
Intermediate answer: No

Follow up: What is the broad topic category of the claim?
Intermediate answer: Economy.

Follow up: Is the topic category sensitive?
Intermediate answer: Yes

Follow up: Can the claim be harmful if false?
Intermediate answer: Yes

So the final answer is: Yes

Question: Classify if the sentences below are check-worthy: [sentences]

Answer in binary Yes or No. \\
\hline
\end{tabular}
\caption{HCoT prompt template.}
\label{tab:prompt-hcot}
\vspace{-10pt}
\end{table}

\begin{table}[ht!]
\centering
\small
\begin{tabular}{|p{0.45\textwidth}|}
\hline
\textbf{User message:} Step 1: Check-worthiness of sentences. Sentences are in 16 different natural languages. We need to determine if the sentences contain a claim that a professional fact-checker should verify. To do this, we'll consider whether the claim is likely to be false, is of public interest, and/or appears to be harmful. What are our thoughts on this?

Step 2: Evaluating False Information. Now, let's think about the likelihood of false information in the sentence. On a scale of 1-5, where 1 is "NO, definitely contains no false information" and 5 is "YES, definitely contains false information", what's our assessment?

Step 3: Considering Public Interest. Next, we need to evaluate whether the sentence's claim has an impact on or is of interest to the general public. Using the same 1-5 scale, where 1 is "NO, definitely not of interest" and 5 is "YES, definitely of interest", what do we think?

Step 4: Assessing Harmfulness. Now, let's consider the potential harm caused by the sentence. On the same 1-5 scale, where 1 is "NO, definitely not harmful" and 5 is "YES, definitely harmful", what's our evaluation?

Step 5: Conclusion. Taking into account our thoughts from the previous steps, do we think the sentence is a "claim" that requires verification or a "non-claim" that doesn't require further investigation?

[sentence]

Our final answer is, do not display any explanations or intermediate steps: \\
\hline
\end{tabular}
\caption{CoT\_CLEF\_2N prompt template.}
\label{tab:prompt-cot-clef-2n}
\vspace{-15pt}
\end{table}

In all evaluations, model outputs were strictly constrained to binary judgments in order to ensure comparability across prompts and systems. Whenever the model produced an ambiguous or non-binary response (e.g., free-form text, justification without a decision, or mixed labels), we employed a short fallback clarification prompt that explicitly enforced a \texttt{Yes}/\texttt{No} decision. This procedure allowed us to maintain consistency in the labeling process while still benefiting from the richer reasoning traces generated under the CoT variants.

The six prompt designs represent a spectrum of prompting strategies, ranging from minimal instructions (GA) to structured multi-step reasoning with intermediate dimensions (CoT-style prompts). Their inclusion in the evaluation was intended to (i) probe the sensitivity of LLMs to prompt design, (ii) measure the impact of guided reasoning versus concise instructions, and (iii) assess generalizability in multilingual settings.

Empirically, we found that the CoT\_CLEF\_on\_Q prompt increased interpretability of the outputs but occasionally led to verbose or partially off-task answers, hence the necessity of the fallback clarification step. The combination of these prompt variants and the fallback mechanism provided a robust evaluation framework, ensuring both fairness across models and insight into the trade-offs between prompt complexity, interpretability, and reliability.

\section{Topic detection}
\label{sec:appendix-topic-detection}
In this section, we present the process of Topic detection in the final dataset described in Section~\ref{sec:dataset_statistics}. Since the source datasets provide only general information about their origin, no per-sample topic annotations are available. To address this, we performed topic detection for each sample in the final dataset using the Llama3:4B model and the prompt described in ~\ref{tab:topic_detection_prompt}, classifying samples into seven predefined categories (\textit{health}, \textit{politics}, \textit{environment}, \textit{science}, \textit{sport}, \textit{entertainment}, \textit{history}). Samples that could not be confidently assigned to a single category were labeled as \textit{various}, representing mixed or ambiguous topics. The resulting topic distributions for the in-distribution and OOD datasets are reported in Tables~\ref{tab:multicw_topics} and~\ref{tab:ood_topics}, and are further illustrated in Figures~\ref{fig:topic-distribution-multicw} and~\ref{fig:topic-distribution-ood}.

\begin{table}[t]
\centering
\caption{Topic distribution in the MultiCW dataset}
\label{tab:multicw_topics}
\begin{tabular}{l c}
\hline
\textbf{Topic} & \textbf{Percentage (\%)} \\
\hline
Health         & 20.04 \\
Politics       & 24.22 \\
Environment    & 2.12  \\
Science        & 2.42  \\
Sport          & 0.61  \\
Entertainment  & 3.20  \\
History        & 0.06  \\
Various        & 21.95 \\
\hline
\end{tabular}
\end{table}

\begin{table}[ht!]
\centering
\small
\begin{tabular}{|p{0.45\textwidth}|}
\hline
\textbf{Prompt:}
prompt = "Classify the following text into one of the categories: [health, politics, environment, science, sport, entertainment]. Respond in English and provide strictly one category from the list, without any additional commentary. If the text does not match any category, respond with 'variable."
['health', 'politics', 'environment', 'science', 'sport', 'entertainment', 'history'] \\
\hline
\end{tabular}
\caption{Topic detection prompt template.}
\label{tab:topic_detection_prompt}
\end{table}

\begin{table}[t]
\centering
\caption{Topic distribution in the OOD dataset}
\label{tab:ood_topics}
\begin{tabular}{l c}
\hline
\textbf{Topic} & \textbf{Percentage (\%)} \\
\hline
Health         & 14.49 \\
Politics       & 39.97 \\
Environment    & 5.31  \\
Science        & 4.75  \\
Sport          & 1.50  \\
Entertainment  & 7.92  \\
History        & 1.54  \\
Various        & 24.52 \\
\hline
\end{tabular}
\end{table}

\section{Per-language Fine-tuned models evaluation}
\label{sec:appendix-per-language-evaluation}
In this section, we present the per-language evaluation of fine-tuned Transformer models (Section~\ref{tab:transformer_per_language}) and zero-shot performance of large language models (Section~\ref{tab:llms_per_language}). Due to space constraints, we report results for the three best-performing LLMs out of 15 evaluated, alongside all three fine-tuned Transformer models. Additional details regarding the evaluation methodology are provided in Section~\ref{sec:results}.
\vspace{-30pt}

\begin{table}[ht!]
\centering
\footnotesize
\setlength{\tabcolsep}{6pt}
\renewcommand{\arraystretch}{1.0}
\begin{tabular}{lccc}
\hline
\textbf{Metric} & 
\textbf{\shortstack{XLM-R \\ \phantom{CoT}}} & \textbf{\shortstack{mDeBERTa \\ \phantom{CoT}}} & \textbf{\shortstack{LESA \\ \phantom{CoT}}} \\
\hline
\multicolumn{4}{l}{\textbf{Arabic}} \\
Accuracy            & 0.80 & 0.81 & 0.67 \\
Precision (macro)   & 0.82 & 0.82 & 0.70 \\
Recall (macro)      & 0.80 & 0.81 & 0.67 \\
\hline
\multicolumn{4}{l}{\textbf{Bulgarian}} \\
Accuracy            & 0.88 & 0.90 & 0.53 \\
Precision (macro)   & 0.89 & 0.90 & 0.53 \\
Recall (macro)      & 0.88 & 0.90 & 0.53 \\
\hline
\multicolumn{4}{l}{\textbf{Bengali}} \\
Accuracy            & 0.97 & 0.96 & 0.82 \\
Precision (macro)   & 0.97 & 0.96 & 0.84 \\
Recall (macro)      & 0.97 & 0.96 & 0.82 \\
\hline
\multicolumn{4}{l}{\textbf{Czech}} \\
Accuracy            & 0.91 & 0.90 & 0.78 \\
Precision (macro)   & 0.91 & 0.91 & 0.78 \\
Recall (macro)      & 0.91 & 0.90 & 0.78 \\
\hline
\multicolumn{4}{l}{\textbf{German}} \\
Accuracy            & 0.96 & 0.97 & 0.88 \\
Precision (macro)   & 0.96 & 0.97 & 0.88 \\
Recall (macro)      & 0.96 & 0.97 & 0.88 \\
\hline
\multicolumn{4}{l}{\textbf{English}} \\
Accuracy            & 0.88 & 0.88 & 0.68 \\
Precision (macro)   & 0.88 & 0.88 & 0.68 \\
Recall (macro)      & 0.87 & 0.88 & 0.68 \\
\hline
\multicolumn{4}{l}{\textbf{Spanish}} \\
Accuracy            & 0.88 & 0.88 & 0.76 \\
Precision (macro)   & 0.89 & 0.88 & 0.78 \\
Recall (macro)      & 0.88 & 0.88 & 0.76 \\
\hline
\multicolumn{4}{l}{\textbf{French}} \\
Accuracy            & 0.96 & 0.97 & 0.88 \\
Precision (macro)   & 0.96 & 0.97 & 0.89 \\
Recall (macro)      & 0.96 & 0.97 & 0.88 \\
\hline
\multicolumn{4}{l}{\textbf{Hindi}} \\
Accuracy            & 0.98 & 0.99 & 0.87 \\
Precision (macro)   & 0.98 & 0.99 & 0.88 \\
Recall (macro)      & 0.98 & 0.99 & 0.87 \\
\hline
\multicolumn{4}{l}{\textbf{Polish}} \\
Accuracy            & 0.95 & 0.94 & 0.83 \\
Precision (macro)   & 0.95 & 0.94 & 0.83 \\
Recall (macro)      & 0.95 & 0.94 & 0.83 \\
\hline
\multicolumn{4}{l}{\textbf{Portuguese}} \\
Accuracy            & 0.98 & 0.98 & 0.91 \\
Precision (macro)   & 0.98 & 0.98 & 0.92 \\
Recall (macro)      & 0.98 & 0.98 & 0.91 \\
\hline
\multicolumn{4}{l}{\textbf{Russian}} \\
Accuracy            & 0.93 & 0.94 & 0.86 \\
Precision (macro)   & 0.93 & 0.94 & 0.86 \\
Recall (macro)      & 0.93 & 0.94 & 0.86 \\
\hline
\multicolumn{4}{l}{\textbf{Slovak}} \\
Accuracy            & 0.90 & 0.91 & 0.78 \\
Precision (macro)   & 0.91 & 0.91 & 0.78 \\
Recall (macro)      & 0.90 & 0.91 & 0.78 \\
\hline
\multicolumn{4}{l}{\textbf{Turkish}} \\
Accuracy            & 0.88 & 0.90 & 0.68 \\
Precision (macro)   & 0.88 & 0.90 & 0.71 \\
Recall (macro)      & 0.88 & 0.90 & 0.68 \\
\hline
\multicolumn{4}{l}{\textbf{Ukrainian}} \\
Accuracy            & 0.92 & 0.92 & 0.81 \\
Precision (macro)   & 0.92 & 0.92 & 0.81 \\
Recall (macro)      & 0.92 & 0.92 & 0.81 \\
\hline
\multicolumn{4}{l}{\textbf{Chinese}} \\
Accuracy            & 0.92 & 0.92 & 0.84 \\
Precision (macro)   & 0.92 & 0.92 & 0.85 \\
Recall (macro)      & 0.92 & 0.92 & 0.84 \\
\hline
\end{tabular}
\caption{Per language performance of fine-tuned Transformer models on the MultiCW test set. Results are reported as accuracy, macro-precision, and macro-recall per each language.}
\vspace{-10pt}
\label{tab:transformer_per_language}
\end{table}

\begin{table}[ht!]
\centering
\footnotesize
\setlength{\tabcolsep}{6pt}
\renewcommand{\arraystretch}{1.0}
\begin{tabular}{lccc}
\hline
\textbf{Metric} & \textbf{\shortstack{Nemotron-4\\340B (CoT)}} & \textbf{\shortstack{Claude 3.5\\Haiku (CoT)}} & \textbf{\shortstack{Llama 3.1\\70B (CoT)}} \\
\hline
\multicolumn{4}{l}{\textbf{Arabic}} \\
Accuracy        & 0.64 & 0.60 & 0.58 \\
Precision (macro) & 0.67 & 0.61 & 0.60 \\
Recall (macro)    & 0.64 & 0.60 & 0.58 \\
\hline
\multicolumn{4}{l}{\textbf{Bulgarian}} \\
Accuracy        & 0.66 & 0.59 & 0.54 \\
Precision (macro) & 0.66 & 0.59 & 0.54 \\
Recall (macro)    & 0.66 & 0.59 & 0.54 \\
\hline
\multicolumn{4}{l}{\textbf{Bengali}} \\
Accuracy        & 0.81 & 0.79 & 0.74 \\
Precision (macro) & 0.81 & 0.80 & 0.74 \\
Recall (macro)    & 0.81 & 0.79 & 0.74 \\
\hline
\multicolumn{4}{l}{\textbf{Czech}} \\
Accuracy        & 0.81 & 0.75 & 0.77 \\
Precision (macro) & 0.81 & 0.75 & 0.77 \\
Recall (macro)    & 0.81 & 0.75 & 0.77 \\
\hline
\multicolumn{4}{l}{\textbf{German}} \\
Accuracy        & 0.87 & 0.82 & 0.81 \\
Precision (macro) & 0.87 & 0.82 & 0.81 \\
Recall (macro)    & 0.87 & 0.82 & 0.81 \\
\hline
\multicolumn{4}{l}{\textbf{English}} \\
Accuracy        & 0.76 & 0.61 & 0.65 \\
Precision (macro) & 0.76 & 0.61 & 0.66 \\
Recall (macro)    & 0.76 & 0.61 & 0.65 \\
\hline
\multicolumn{4}{l}{\textbf{Spanish}} \\
Accuracy        & 0.74 & 0.74 & 0.73 \\
Precision (macro) & 0.77 & 0.74 & 0.74 \\
Recall (macro)    & 0.74 & 0.74 & 0.73 \\
\hline
\multicolumn{4}{l}{\textbf{French}} \\
Accuracy        & 0.85 & 0.80 & 0.79 \\
Precision (macro) & 0.85 & 0.80 & 0.79 \\
Recall (macro)    & 0.85 & 0.80 & 0.79 \\
\hline
\multicolumn{4}{l}{\textbf{Hindi}} \\
Accuracy        & 0.85 & 0.82 & 0.80 \\
Precision (macro) & 0.86 & 0.82 & 0.81 \\
Recall (macro)    & 0.85 & 0.82 & 0.80 \\
\hline
\multicolumn{4}{l}{\textbf{Polish}} \\
Accuracy        & 0.87 & 0.85 & 0.83 \\
Precision (macro) & 0.87 & 0.85 & 0.83 \\
Recall (macro)    & 0.87 & 0.85 & 0.83 \\
\hline
\multicolumn{4}{l}{\textbf{Portuguese}} \\
Accuracy        & 0.82 & 0.80 & 0.79 \\
Precision (macro) & 0.82 & 0.80 & 0.79 \\
Recall (macro)    & 0.82 & 0.80 & 0.79 \\
\hline
\multicolumn{4}{l}{\textbf{Russian}} \\
Accuracy        & 0.86 & 0.84 & 0.85 \\
Precision (macro) & 0.87 & 0.84 & 0.85 \\
Recall (macro)    & 0.86 & 0.84 & 0.85 \\
\hline
\multicolumn{4}{l}{\textbf{Slovak}} \\
Accuracy        & 0.84 & 0.81 & 0.75 \\
Precision (macro) & 0.84 & 0.81 & 0.75 \\
Recall (macro)    & 0.84 & 0.81 & 0.75 \\
\hline
\multicolumn{4}{l}{\textbf{Turkish}} \\
Accuracy        & 0.70 & 0.67 & 0.64 \\
Precision (macro) & 0.75 & 0.68 & 0.67 \\
Recall (macro)    & 0.70 & 0.67 & 0.64 \\
\hline
\multicolumn{4}{l}{\textbf{Ukrainian}} \\
Accuracy        & 0.83 & 0.79 & 0.80 \\
Precision (macro) & 0.84 & 0.79 & 0.80 \\
Recall (macro)    & 0.83 & 0.79 & 0.80 \\
\hline
\multicolumn{4}{l}{\textbf{Chinese}} \\
Accuracy        & 0.79 & 0.83 & 0.80 \\
Precision (macro) & 0.80 & 0.83 & 0.80 \\
Recall (macro)    & 0.79 & 0.83 & 0.80 \\
\hline
\end{tabular}
\caption{Zero-shot performance of the top three large language models from our benchmark on the MultiCW test set, reported per language. Accuracy, macro precision, and macro recall are reported for each language."}
\vspace{-10pt}
\label{tab:llms_per_language}
\end{table}

\hfill \break

\section{Training Configuration}
\label{sec:appendix-configuration}
Fine-tuning was conducted using the \texttt{keras-hub} implementations of 
\texttt{XLM-RoBERTaTextClassifier} and \texttt{DeBERTaV3Classifier}, and a custom \texttt{LESAClaimModel}. 
For all models, we set the maximum input sequence length to 256 tokens (60 tokens in LESA’s BERT encoder) 
and used a batch size of 32. 
All layers were unfrozen during fine-tuning.  

For \textbf{XLM-R}, we trained using Adam with a constant learning rate of $2\cdot 10^{-6}$ for 5 epochs.  
For \textbf{mDeBERTa}, we used Adam with a cosine decay schedule, 
starting at $2\cdot 10^{-6}$ and decaying to 50\% of the initial rate over 5 epochs.  
For \textbf{LESA}, we trained the semantic modules and BERT encoder jointly for 5 epochs, 
using Adam with a learning rate of $2\cdot 10^{-6}$.  
Dropout of 0.2 was applied in all models. 
Models were trained on a VM on Azure platform with NVIDIA A100 PCIe GPU with 80 GB memory.

\section{Use of Generative Models}

During the writing of this paper, we have utilized the large language models in order to improve the grammar, sentence structure and the overall flow of some sections. In all cases, the texts were first written by the authors, passed to the LLM for improvement and carefully checked afterwards in order to check the meaning of the sentence has not changed.

\end{document}